\documentclass[lettersize,journal]{IEEEtran}
\usepackage{amsmath,amsfonts}
\usepackage{booktabs}
\usepackage{algorithmic}
\usepackage{algorithm}
\usepackage{array}
\usepackage[caption=false,font=normalsize,labelfont=sf,textfont=sf]{subfig}
\usepackage{textcomp}
\usepackage{stfloats}
\usepackage{url}
\usepackage{xurl}
\usepackage{verbatim}
\usepackage{graphicx}
\usepackage{cite}
\usepackage{balance}
\usepackage{amssymb}

\begin{document}

\title{A Content-Aware Pure Permutation with Intrinsic Avalanche Effect: Breaking the Diffusion-Permutation Dichotomy}

\author{Zahra Ghoraeian,
        Mohammad-Reza Sadeghi,
        and Samaneh Mashhadi,
\thanks{Z. Ghoraeian is with the Department of Mathematics and Computer Science, Amirkabir University of Technology (Tehran Polytechnic), Tehran, Iran (e-mail: zahra.ghoraeian@aut.ac.ir). ORCID: 0009-0009-3547-4437.}
\thanks{M.-R. Sadeghi is with the Department of Mathematics and Computer Science, Amirkabir University of Technology (Tehran Polytechnic), Tehran, Iran (e-mail: msadeghi@aut.ac.ir). ORCID: 0000-0002-7676-4168.}
\thanks{S. Mashhadi is with the School of Mathematics and Computer Science, Iran University of Science and Technology, Narmak, Tehran, 1684613114, Iran (e-mail: smashhadi@iust.ac.ir). ORCID: 0000-0001-9191-1376.}
\thanks{Corresponding author: Mohammad-Reza Sadeghi (e-mail: msadeghi@aut.ac.ir).}
}
\markboth{arXiv}%
{Ghoraeian \MakeLowercase{\textit{et al.}}: TCA: Triangular Content-Aware Permutation algorithm}

\maketitle

\begin{abstract}

Pixel permutation is a fundamental tool in image processing, image encryption, and data hiding (including watermarking and steganography) that transforms spatial arrangement solely by relocating pixels without altering their numerical values. A common assumption in the technical literature holds that permutation alone cannot create differential sensitivity; that is, changing a single pixel in the input, after applying permutation, merely relocates that same pixel in the output, producing no avalanche effect. This paper challenges this assumption by introducing the Triangular Content-Aware Permutation (TCA) algorithm.

The proposed method creates a unique image partitioning by extracting edge points using the Canny operator and applying Delaunay Triangulation to a set comprising edges and corners, then uses this partitioning to construct the permutation pattern. Since the triangulation structure is highly sensitive to image geometry and topology, changing even a single pixel alters the edge map, resulting in a fundamentally different triangulation and an entirely new global permutation pattern. Unlike classical dimension-based permutations-which are still currently used in the permutation stage of data hiding systems-and advanced content-aware modern permutations (2025--2026) employed in the permutation stage of image encryption systems, all of which lack differential sensitivity at this stage-TCA dramatically increases NPCR from near-zero values (in static-formula permutations) to 97.10\% solely through pixel relocation.

Experimental results on 50 standard images demonstrate that TCA, with an average of 14.81 iterations, achieves NPCR = 97.10\% and UACI = 20.06\%, proving that a pure permutation can exhibit significant differential sensitivity. In contrast, conventional methods maintain near-zero NPCR under identical conditions. The number of iterations required to reach the chaos threshold varies from 6.4 to 30.7 depending on content complexity, rooted in deep geometric dependency that renders the algorithm's behavior unpredictable. Low PSNR values (11.93 dB) and near-zero correlation coefficients ($\sim$10\textsuperscript{-3}) confirm the algorithm's superior statistical performance between two images initially differing by only one pixel. Although execution speed is slower than classical methods due to triangulation complexity, this represents a deliberate trade-off between speed and superior qualitative security. Given the non-analytic and content-dependent nature of the permutation pattern-which makes reconstructing the mapping impossible without access to the reference image-the algorithm is well-suited for applications in reference-based encryption, fragile watermarking, and non-blind steganography.
\end{abstract}



\section{Introduction}

\IEEEPARstart{T}{he} increasing use of images in sensitive domains such as healthcare, surveillance, and military communications has drawn widespread attention to the security of visual data \cite{ref1,ref2}. Medical, satellite, and personal images all contain valuable information that requires protection during storage and transmission \cite{ref3,ref4}, consequently making the assurance of security, integrity, and confidentiality of image data a critical area of research \cite{ref5,ref6}.

Many image processing and security applications-including encryption, digital watermarking, steganography, content authentication, and secure image transmission-rely on pixel permutation as a fundamental operation \cite{ref1,ref2,ref3,ref7,ref8,ref9}. This operation rearranges the spatial arrangement of pixels without altering their numerical values, thereby breaking spatial correlations. For instance, in encryption it acts as a confusion layer for decorrelating pixels \cite{ref1,ref2,ref10,ref11,ref6}; in robust watermarking it disperses the watermark to enhance resilience against attacks \cite{ref3}; in steganography it conceals the presence of hidden data \cite{ref7}; and in authentication and fragile watermarking it increases tampering sensitivity by distributing authentication data \cite{ref3}.

Traditional permutation algorithms (Arnold, Zigzag, Spiral) continue to be used in recent research (2024--2025) due to their simplicity and efficiency in hybrid methods \cite{ref4,ref5,ref6,ref7,ref8}. However, these methods suffer from a fundamental limitation: they are dimension-dependent and content-independent, thus generating identical permutation patterns for all images and rendering them vulnerable to KPA and statistical analysis \cite{ref2}. Furthermore, results from Section~\ref{sec:results} show that even after 100 iterations, changing a single pixel yields NPCR $\approx$ 0.0004\% and PSNR $>$ 100 dB, indicating the absence of differential sensitivity and avalanche effect in content-independent permutations. Recently, content-aware methods such as block permutation with feedback \cite{ref1} or adaptive scrambling \cite{ref10} have been proposed, but they either involve explicit diffusion or modify pixel values, thus departing from pure permutation. A review of advanced schemes (2024--2026) in Table~\ref{tab:comparison} shows that none have evaluated or claimed an avalanche effect in the permutation stage alone \cite{ref1,ref2,ref6,ref10,ref11,ref13,ref14,ref15,ref16,ref17}.

\setlength{\arrayrulewidth}{0.8pt}
\begin{table*}[!t]
\centering
\footnotesize
\caption{Investigation of the permutation phase's role in creating the avalanche effect in advanced image encryption algorithms (2024--2026)}
\label{tab:comparison}
\begin{tabular}{|c|p{5cm}|p{3.8cm}|c|c|}
\hline
\textbf{Ref. (Year)} & \textbf{Algorithm / Approach} & \textbf{Architecture} & \shortstack{\textbf{Separate}\\ \textbf{Diffusion Phase?}} & \shortstack{\textbf{Avalanche Effect}\\ \textbf{in Permutation Phase}\\ \textbf{Investigated/Claimed?}} \\
\hline
{[1] (2025)} & MIE-SPD (Simultaneous Permutation-Diffusion) & Multi-image encryption with SPD & $\checkmark$ Simultaneous & $\times$ \\
\hline
{[2] (2024)} & Block permutation + weighted bit-plane chain diffusion & Two-phase encryption & $\checkmark$ Yes & $\times$ \\
\hline
{[10] (2025)} & Bit-level and pixel-level permutation with improved hyperchaotic system & Three-phase encryption & $\checkmark$ Yes & $\times$ \\
\hline
{[11] (2024)} & 1D-CwS + pixel-level permutation with unique numbers & Medical image encryption & $\checkmark$ Yes & $\times$ \\
\hline
{[6] (2024)} & Block permutation + LIS with 2D-HSCM map & Color image encryption & $\checkmark$ Yes & $\times$ \\
\hline
{[13] (2024)} & Dynamic Josephus scrambling + cross-diffusion & Medical image encryption & $\checkmark$ Yes & $\times$ \\
\hline
{[14] (2026)} & SCD-CHAOS (Dynamic S-box + Tent and Henon maps) & Hybrid encryption & $\checkmark$ Yes & $\times$ \\
\hline
{[15] (2025)} & 2D-SQICS + random area permutation & Encryption with new hyperchaotic system & $\checkmark$ Yes & $\times$ \\
\hline
{[16] (2026)} & SQMCML + random-trajectory Josephus permutation & Encryption with CML and CA & $\checkmark$ Yes & $\times$ \\
\hline
{[17] (2026)} & 1D-PCQM + dynamic sequence selection (D3CM-IES) & Encryption with powered Chebyshev map & $\checkmark$ Yes & $\times$ \\
\hline
\textbf{[Proposed] (2026)} & \textbf{Triangular Content-Aware Permutation} & \textbf{Content-dependent pure permutation} & $\times$ No & $\checkmark$ \textbf{Yes (Intrinsic)} \\
\hline
\end{tabular}
\end{table*}

As observed in Table~\ref{tab:comparison}, none of the advanced methods published in recent years (2024--2026) claim differential sensitivity or avalanche effect for their permutation layer. For instance, the content-aware scrambling method in \cite{ref2} enters the diffusion domain by modifying pixel values. Content-independent permutations also produce fixed patterns for identical dimensions \cite{ref8,ref12} and remain vulnerable to statistical and structural attacks \cite{ref2}. Our experimental results show near-zero NPCR (0.0004\%) for Arnold transform as a representative of content-independent permutations. The TCA algorithm, by introducing a dynamic content-aware permutation, provides a unique dispersion map based on the image's topological features and, for the first time, realizes an intrinsic avalanche effect within a pure permutation layer.

\subsection{Proposed Method}

This paper challenges the assumption that permutation alone cannot create differential sensitivity by introducing the Triangular Content-Aware Permutation (TCA) algorithm. The key idea is to make the permutation pattern dependent on the image topology, so that a change in a single pixel transforms the entire structure. TCA achieves this in three steps: edge extraction with Canny, Delaunay triangulation of points, and pixel rearrangement based on triangle traversal. Since Delaunay triangulation depends on point geometry, any image change completely alters the permutation pattern. This geometric dependency creates a unique, analytically unreconstructable pattern, making pattern recovery impossible without the reference image and rendering the algorithm ideal for fragile watermarking and non-blind steganography.

The main contributions of this work are:

\begin{enumerate}
\item \textbf{Conceptual Innovation:} Introduction of the first content-dependent permutation function that, unlike content-independent methods, generates unique and unpredictable permutation patterns for each image and, unlike both classical and modern methods, possesses avalanche effect and differential sensitivity.

\item \textbf{Quantitative Validation:} We demonstrate that TCA achieves NPCR = 97.1\% with an average of 14.81 iterations-a value currently assumed to be near zero for permutations. Although the computational cost of TCA to reach this level of differential sensitivity (4.54 seconds for $512 \times 512$ images, equivalent to approximately 15 iterations) is higher, this cost is fully justified by achieving unparalleled security features such as KPA resistance and over 97\% improvement in permutation differential sensitivity. It is worth noting that the number of iterations is tunable depending on the user's requirements; for instance, only 5 iterations (1.49 seconds) are sufficient to break spatial correlation, and even a single iteration (0.298 seconds) suffices to guarantee the uniqueness of the permutation pattern for each image. Moreover, for non-real-time applications (such as watermarking and authentication) where security takes precedence over speed, this cost is acceptable. It should be noted that TCA is the first algorithm to introduce a pure permutation with intrinsic avalanche effect, challenging a long-standing assumption in the technical literature. As a pioneering work, it is not expected to achieve absolute optimality in all dimensions; rather, its value lies in opening a new path and proving the feasibility of what was previously considered impossible. This paper provides a foundation for future research that will further optimize execution speed.

\item \textbf{Comprehensive Security Analysis:} We provide a complete evaluation including correlation analysis, differential attack resistance, and Known-Plaintext Attack (KPA) analysis, confirming that TCA effectively breaks pixel correlations and resists KPA.
\end{enumerate}

The remainder of this paper is organized as follows: Section~\ref{sec:background} presents the theoretical foundations, Section~\ref{sec:related} reviews related work, Section~\ref{sec:method} details the proposed TCA algorithm, Section~\ref{sec:results} presents experimental results, Section~\ref{sec:limitations} discusses limitations and future work, and Section~\ref{sec:conclusion} concludes the paper.


\section{Theoretical Foundations}\label{sec:background}

The proposed algorithm is designed based on the integration of several fundamental concepts in image processing and data hiding. In this section, we first introduce traditional scrambling methods that are still widely used in watermarking and steganography applications, serving as a background to demonstrate the existing limitations. We then examine the foundations of edge extraction and Delaunay triangulation as the core components of the new method. The goal is to provide the necessary theoretical framework for a better understanding of the proposed scheme's mechanism.

\subsection{Common Image Pixel Permutation Algorithms in Watermarking and Steganography}

Image scrambling is a key step in watermarking and steganography algorithms. Three common algorithms in this field are the Arnold, Spiral, and Zigzag transforms.

\subsubsection{Arnold Transform}

The Arnold transform is one of the mappings used for image pixel permutation. The 2D Arnold map moves a pixel at coordinates $(x, y)$ to new coordinates $(x', y')$ using the following equation, where $N$ is the image dimension. This map is reversible and applicable only to square images.

\begin{equation}
\label{eq:arnold}
\begin{bmatrix}
x' \\
y'
\end{bmatrix}
=
\begin{bmatrix}
1 & 1 \\
1 & 2
\end{bmatrix}
\begin{bmatrix}
x \\
y
\end{bmatrix}
\pmod{N}
\end{equation}

This map depends solely on the image dimensions and acts identically on all images of the same size. An important characteristic of this map is its periodicity, meaning that after a finite number of iterations, the image returns to its original state \cite{ref18,ref19}.

\subsubsection{Zigzag Transform}

\begin{figure}[!t]
\centering
\includegraphics[width=2.5in]{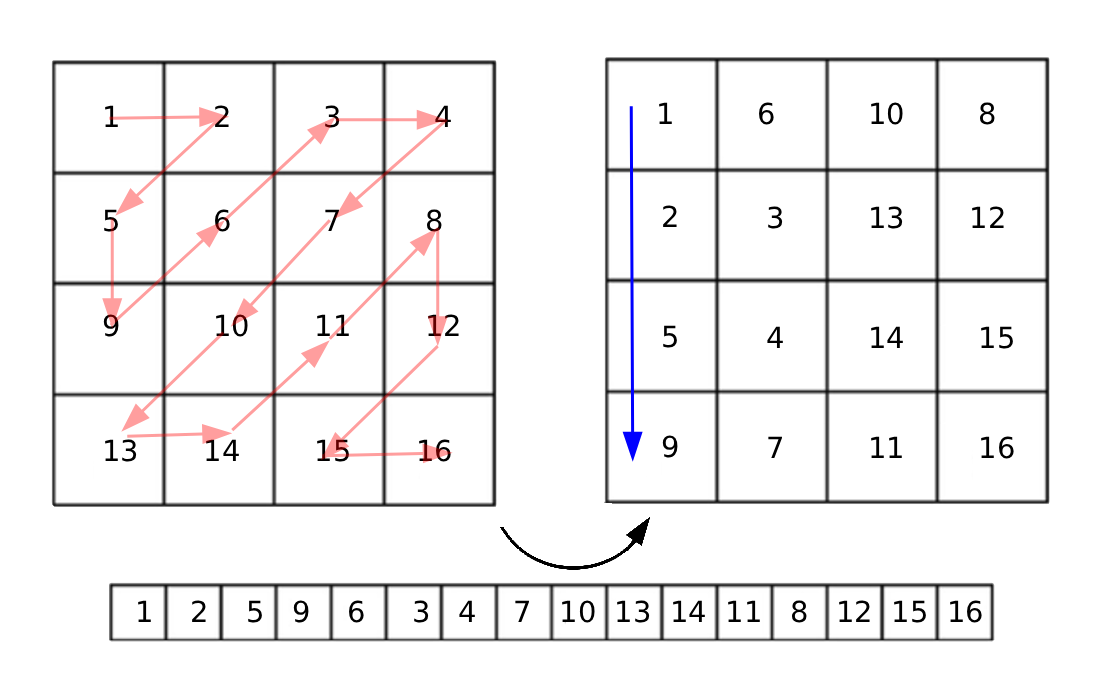}
\caption{Zigzag scrambling pattern}
\label{fig:zigzag}
\end{figure}

In the Zigzag transform, image pixels are scanned sequentially based on a ``Z''-shaped path. This method is also only defined for square images and is reversible. The scrambling pattern generated by this transform is independent of the image content and depends only on its dimensions \cite{ref20}. Figure~\ref{fig:zigzag} shows an example of this scanning pattern.

\subsubsection{Spiral Transform}

\begin{figure}[!t]
\centering
\includegraphics[width=2.5in]{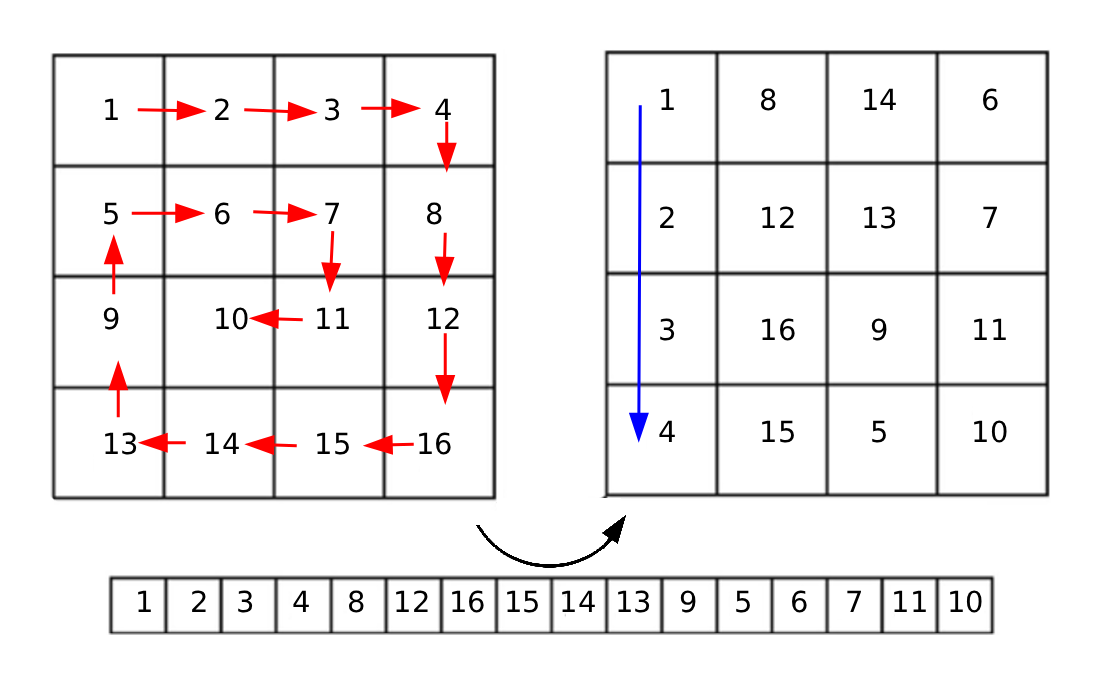}
\caption{Spiral scrambling pattern}
\label{fig:spiral}
\end{figure}

The Spiral transform scrambles the image by scanning pixels from a starting point (usually a corner of the image) in a specific direction. This method is also limited to square images, and its pattern is identical and predictable for same-sized images \cite{ref12}. Figure~\ref{fig:spiral} depicts an example of the spiral scanning pattern.

A critical criterion for evaluating the security of these algorithms is their resistance to Known-Plaintext Attacks (KPA). Since the scrambling pattern in Arnold, Zigzag, and Spiral transforms depends only on the image dimensions and is the same for all same-sized images, if this map is revealed for one image, the security of all other images of that size is completely compromised. This inherent weakness exacerbates the vulnerability of these algorithms to this attack.

As observed, conventional scrambling algorithms suffer from common limitations, the most important of which is their exclusive reliance on image dimensions and the generation of an identical, predictable pattern for same-sized images. Our proposed algorithm addresses these weaknesses with a novel approach using Delaunay triangulation and edge detection, creating a unique and content-dependent scrambling pattern that significantly increases its security.

\subsection{Foundations of Feature Extraction and Triangulation}

\subsubsection{Image Edges and the Canny Algorithm}

Image edges are points with sudden changes in brightness intensity that define the boundaries between different regions. These points are used as key features for image processing. The Canny algorithm is a common method for extracting edges with the following steps:

\begin{enumerate}
\item Applying a Gaussian filter to reduce noise.
\item Calculating the image gradient to identify points with sharp changes.
\item Performing Non-Maximum Suppression to thin the edges.
\item Applying double thresholding to separate strong edges from weak ones.
\end{enumerate}

\begin{figure}[!t]
\centering
\includegraphics[width=2.5in]{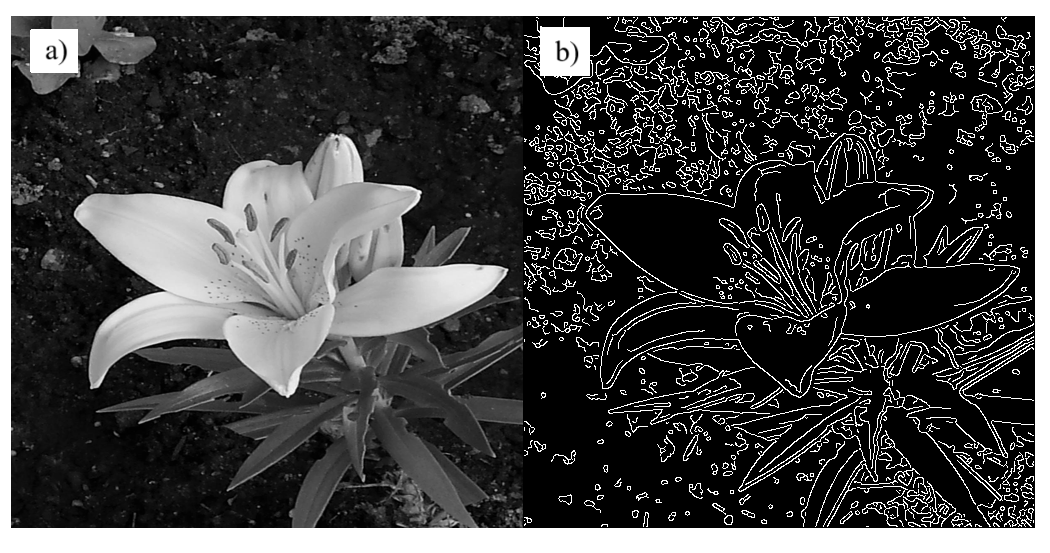}
\caption{(a) Original image; (b) Edge points extracted by Canny algorithm}
\label{fig:canny}
\end{figure}

The edge points extracted by the Canny algorithm are used as the initial input for the Delaunay triangulation process to create a content-dependent triangular mesh for the image. Figure~\ref{fig:canny} shows an example of an image (a) and its edge points extracted by the Canny algorithm (b) \cite{ref21}.

\subsubsection{Delaunay Triangulation (DT)}

Triangulation is a fundamental technique in computational geometry. Delaunay Triangulation (DT) is an optimal partitioning method that divides the convex hull of points into non-overlapping triangles. This method is based on the ``empty circumcircle criterion'' (no other point exists inside the circumcircle of any triangle). This unique property ensures that the internal angles of the triangles are maximized, preventing the formation of long, narrow triangles. The vital point for the TCA algorithm is that the partitioning nature of DT guarantees that every image pixel belongs to exactly one region \cite{ref22}.

\begin{figure*}[!t]
\centering
\includegraphics[width=7.2in]{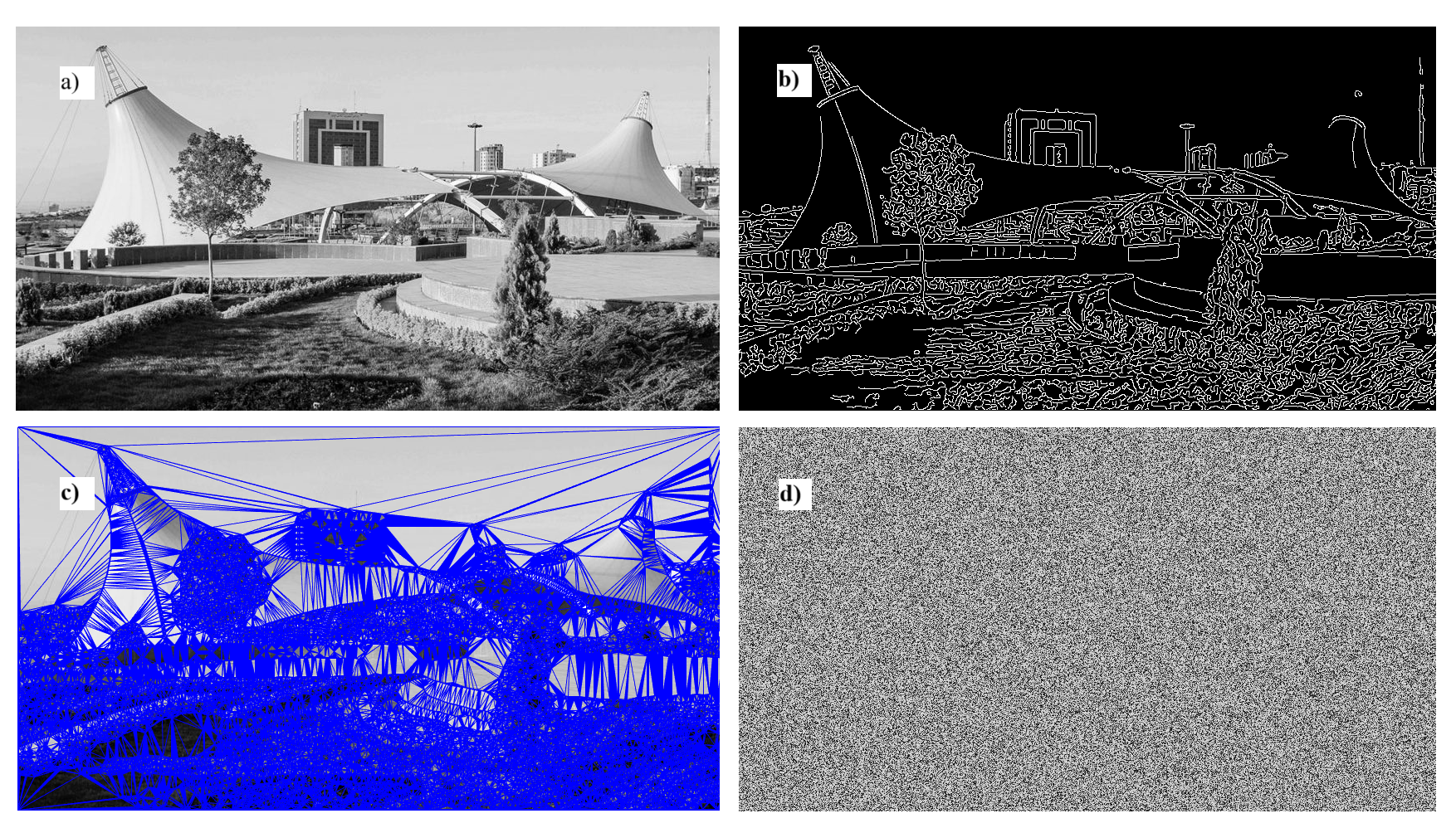}
\caption{Visual example of the TCA algorithm's stages on a $463 \times 858$ pixel image. (a) Original image. (b) Image after edge detection and key point identification. (c) Image with Delaunay triangulation on key points. (d) Final scrambled image after 10 algorithm iterations.}
\label{fig:visual}
\end{figure*}

\subsection{Scrambling Evaluation Metrics}

Various metrics are used to quantitatively evaluate the performance of a scrambling algorithm. A successful scrambling algorithm is characterized by the following criteria:

\begin{itemize}
\item \textbf{Low PSNR and SSIM:} Indicating a severe lack of visual and structural similarity between the original and scrambled images \cite{ref24,ref23}.
\item \textbf{High MSE and BER:} Indicating extensive changes at the numerical and bit levels \cite{ref24}.
\item \textbf{Low Absolute NCC and Autocorrelation (AC):} Showing the successful elimination of any linear statistical relationship and spatial correlation between neighboring pixels, respectively \cite{ref23,ref24}.
\end{itemize}

\subsubsection{Differential Sensitivity Metrics: NPCR and UACI}

The standard metrics for measuring differential sensitivity are NPCR (percentage of different pixels) and UACI (average intensity difference), calculated between two images differing by one pixel after applying the algorithm. The theoretical ideal values for these metrics in a complete encryption system are 99.61\% and 33.46\%, respectively \cite{ref23}. However, these values are defined for complete encryption algorithms that include both permutation and diffusion stages.

It is worth noting that content-independent permutations (such as the Arnold transform) inherently lack the avalanche effect, because changing one pixel in the original image only results in the relocation of two pixels in the output image. Consequently, NPCR equals $2/(M \times N)$ and UACI equals $2d/(M \times N \times 255)$ (where $d$ is the pixel intensity difference and $M \times N$ is the image size), which for images of typical dimensions are negligible (near zero) \cite{ref1}. 

Although the common assumption is that permutations in general (even content-aware ones) lack differential sensitivity, the proposed TCA algorithm fundamentally differs from existing permutations. Its iterative and content-dependent mechanism-based on edge points and Delaunay triangulation-creates a powerful dynamic diffusion effect that is absent in existing permutations and leads to differential sensitivity. Therefore, differential metrics like NPCR and UACI are not only applicable but are crucial for quantifying TCA's superior differential resistance against Known-Plaintext Attacks (KPA). This unique behavior elevates TCA beyond the scope of simple existing methods.

\subsection{Entropy and Histogram in the TCA Mechanism}

By definition, permutations only relocate pixels and do not change their values. Therefore, statistical properties such as entropy and histogram remain constant. In TCA, these properties also remain unchanged; hence, analyzing these metrics, unlike encryption algorithms that also modify pixel values, is ineffective.

\section{Related Work}\label{sec:related}

To articulate the research gap that this study intends to fill, the literature review is structured around two main axes: (1) the established permutation-diffusion paradigm and the passive role of permutation in image encryption, and (2) the continued use of classical scramblers in recent publications. Finally, by synthesizing these two axes, the innovation and distinction of the proposed Triangular Content-Aware (TCA) algorithm will be clearly defined.

\subsection{The Permutation-Diffusion Paradigm and the Passive Role of Permutation in Image Encryption}

Since Fridrich's seminal work in 1998 \cite{ref25}, the two-stage permutation-diffusion architecture has become the dominant paradigm in chaos-based image encryption. In the first stage, pixel positions are rearranged to break spatial correlations, and in the second stage, pixel values are modified to achieve the avalanche effect-a phenomenon in complete cryptographic systems where a single-bit change in the input propagates to approximately half of the output bits \cite{ref26,ref27}.

Numerous studies have improved this framework through various innovations, including new chaotic maps \cite{ref28,ref29}, bit-level operations \cite{ref30,ref31}, DNA coding \cite{ref32,ref33}, and plaintext-dependent key generation mechanisms \cite{ref34,ref35}. However, a rigorous review of these schemes reveals a consistent assumption: the avalanche effect is exclusively the domain of the diffusion phase, while permutation is regarded merely as a ``passive layer.''

Table~\ref{tab:comparison} (in the Introduction) provides an overview of prominent and up-to-date algorithms (2024--2026). This review demonstrates that in all investigated schemes-whether they utilize separate diffusion phases \cite{ref26,ref2,ref10,ref11}, employ simultaneous permutation-diffusion operations \cite{ref1}, or even pure permutation methods \cite{ref36}-the permutation layer alone lacks an intrinsic avalanche effect.

It is evident that a purely geometric and content-independent permutation, due to its lack of dependence on pixel information, cannot respond to changes in the input image (such as a single-pixel change) and consequently lacks any differential sensitivity or avalanche effect. In other words, as long as the pixel rearrangement pattern is not a function of the image content, changes in the input will have no effect on the permutation output, and NPCR will remain near zero at best, as confirmed by the examination of the Arnold transform as a representative of content-independent permutations (Table~\ref{tab:results}). These findings align with comparative studies in the field \cite{ref37}, which focus on quality metrics rather than differential security measures. These permutations also lack security against Known-Plaintext Attacks (KPA), because if the permutation pattern is content-independent (i.e., depends only on image dimensions), having one pair (original image, scrambled image) allows an attacker to easily reconstruct the permutation pattern for all images of the same size. Nevertheless, the construction of such permutations with novel ideas is still prevalent \cite{ref6,ref8,ref9,ref13,ref23}.

These fundamental limitations have driven researchers toward designing ``content-aware'' permutations. In this direction, Wen et al. \cite{ref2}, by proposing block permutation based on plaintext feedback, took a step toward making the permutation pattern dependent on image content; however, this method, by modifying pixel values at the bit level, goes beyond the boundaries of pure permutation and effectively enters the diffusion domain. Similarly, Zhang and Zhang \cite{ref10}, by introducing bit-level and pixel-level permutation in an improved hyperchaotic system, attempted to increase permutation complexity, but again deferred the avalanche effect to the bidirectional diffusion phase. In the domain of medical images, Zhang et al. \cite{ref13} utilized dynamic Josephus scrambling, and Yogi et al. \cite{ref14}, by presenting SCD-CHAOS, combined Tent permutation with a dynamic S-box; however, in all these cases, pixel value modification and avalanche effect creation are still performed in a separate diffusion phase. Likewise, Li et al. \cite{ref15} with the 2D-SQICS hyperchaotic system, Wang et al. \cite{ref16} with the SQMCML model and random-trajectory Josephus permutation, and Sarra et al. \cite{ref17} with the 1D-PCQM powered Chebyshev map, have all maintained the permutation-diffusion architecture with a separate diffusion phase. Therefore, despite the remarkable diversity in chaotic sequence generation methods and dynamic permutation mechanisms, none of these approaches have succeeded in creating differential sensitivity within the permutation phase itself, and the avalanche effect remains exclusively within the domain of the diffusion phase. This gap is the primary motivation for presenting the TCA algorithm as the first pure permutation with an intrinsic avalanche effect.

\subsection{Continued Use of Classical Scramblers in Recent Publications}

Despite vulnerability to Known-Plaintext Attacks (KPA) due to their sole dependence on image dimensions, classical scramblers like Arnold, Zigzag, and Spiral transforms remain widely used as the primary basis for permutation construction in recent research (2024--2025) \cite{ref4,ref5,ref6,ref7,ref8,ref23,ref38,ref39,ref40,ref41}; however, all these methods, due to their static nature, produce identical and predictable permutation patterns for all images of the same size and lack differential sensitivity, highlighting the urgent need for a more secure, content-aware alternative.

\subsection{Final Synthesis and Research Gap Statement}

The comprehensive review of related work reveals several critical insights:

\begin{enumerate}
\item \textbf{Passive Permutation Paradigm:} Permutation (confusion) is one of the main stages of image encryption. Yet in current encryption schemes, it is considered a passive layer incapable of creating an avalanche effect. Even advanced algorithms from 2024--2026 \cite{ref1,ref2,ref6,ref10,ref11,ref13,ref14,ref15,ref16,ref17} either rely on a separate diffusion phase or, in the case of pure permutations, exhibit NPCR values near zero.

\item \textbf{Persistence of Vulnerable Classic Methods:} Classical scramblers, despite having static and dimension-dependent formulas and known weaknesses, remain prevalent in recent publications.
\end{enumerate}

\noindent \textbf{Research Gap and TCA Innovation:} It should be noted that the aim of this paper is not to present a complete encryption system, but rather to introduce a novel permutation function applicable to various image security applications-including the permutation stage in encryption, watermarking, and steganography. By intelligently synthesizing these concepts, this paper introduces a new paradigm. The proposed Triangular Content-Aware (TCA) algorithm utilizes the fusion of the Canny edge detector and Delaunay triangulation-not for data embedding or feature extraction, but to generate a unique, unpredictable, and content-dependent pixel permutation map that possesses an intrinsic avalanche effect without any separate diffusion phase.

As shown in Table~\ref{tab:comparison} and our experimental results (Section~\ref{sec:results}), TCA alone, within a pure permutation layer, elevates NPCR from near 0\% to 97.10\%, challenging the long-standing assumption that permutation must be passive. This map serves as an intelligent and secure scrambler that can replace classic methods in watermarking, steganography, and the permutation stage of encryption systems, solving the fundamental problem of KPA vulnerability in these applications. Thus, TCA represents a fundamental shift in conceptualizing ``pixel permutation'' as an active security element.

\section{Proposed Triangular Content-Aware Permutation (TCA) Algorithm}\label{sec:method}

The Triangular Content-Aware Permutation (TCA) algorithm is a novel method for pixel permutation that utilizes image edge points and Delaunay triangulation to generate a unique, non-analytic permutation map for each image. This approach directly addresses the primary vulnerability of conventional algorithms to Known-Plaintext Attacks (KPA) and provides a secure alternative.

\subsection{Algorithm Overview}\label{subsec:tca_overview}
The TCA algorithm operates in three main stages:
\begin{enumerate}
    \item \textbf{Content Feature Extraction (Edge Detection):} The algorithm extracts the content-dependent features of the image, specifically the edge points, using the Canny algorithm.
    \item \textbf{Geometric Partitioning (Delaunay Triangulation):} The extracted edge points are used as the seeds for the Delaunay Triangulation process. This step divides the image plane into a unique set of non-overlapping triangles.
    \item \textbf{Pixel Permutation (Scrambling):} The pixels are then permuted based on the geometric properties of the generated triangles. The unique permutation map is created by traversing the list of triangles and then performing an internal shuffling of the pixels within each triangle region.
\end{enumerate}

\begin{figure*}[h!]
  \centering
  \includegraphics[width=0.55\textwidth]{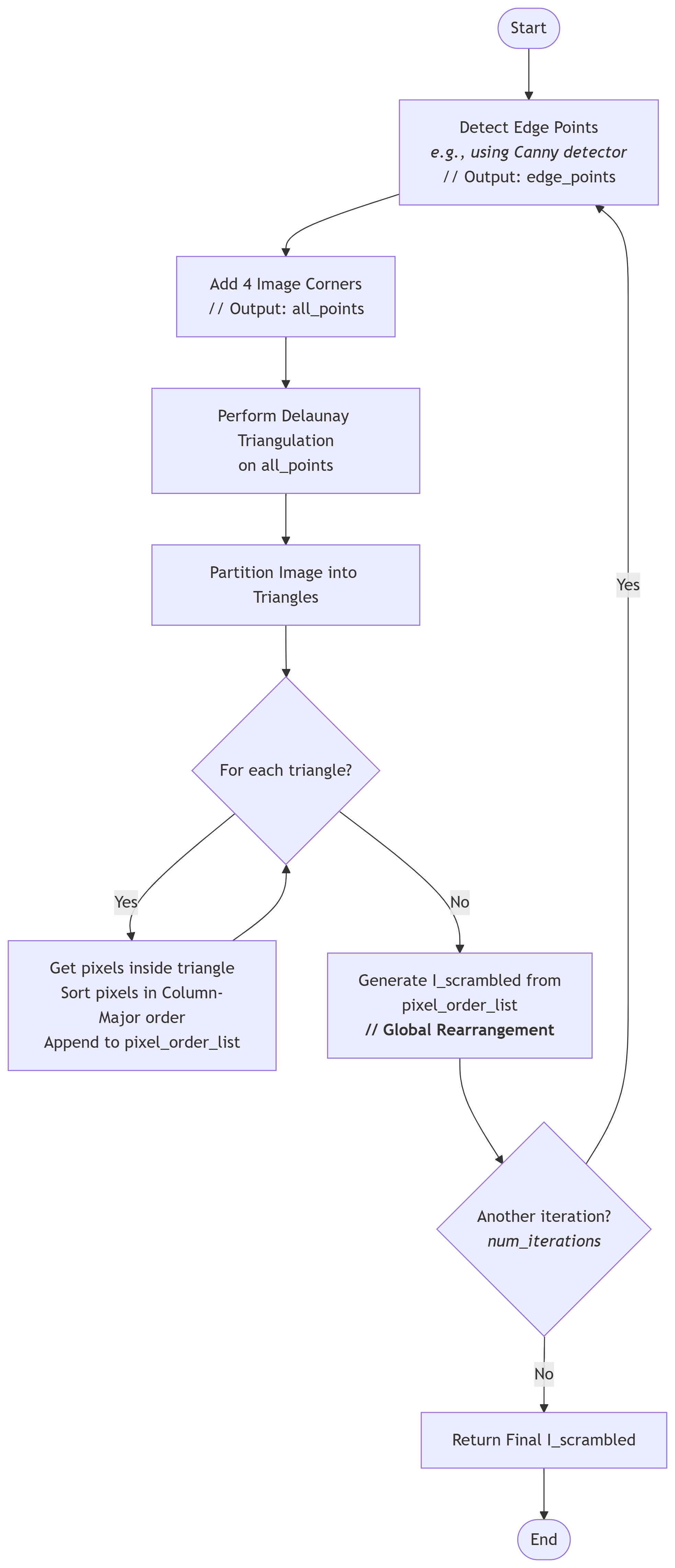}
  \caption{Detailed flowchart of the proposed Triangular Content-Dependent Scrambling Algorithm (TCA)}
 \label{fig:tca_flowchart}
\end{figure*}

The detailed workflow and main mechanism of the proposed algorithm are provided in the comprehensive flowchart in Figure \ref{fig:tca_flowchart} and summarized in Algorithm~\ref{alg:tca}.

\begin{algorithm}[!t]
\caption{ Content-Aware Pure Permutation Algorithm (TCA)}
\label{alg:tca}
\footnotesize
\begin{minipage}{\linewidth}
\begin{algorithmic}[1]
\REQUIRE $I$: Input grayscale image of size $M \times N$
\ENSURE $I_{\text{scrambled}}$: Scrambled image
\STATE $\text{num\_iterations}$: Number of scrambling iterations

\STATE $I_{\text{current}} \gets I$
\FOR{$\text{iter} = 1$ \TO $\text{num\_iterations}$}
    \STATE \textbf{Step 1: Detect edge points}
    \STATE $\text{edge\_points} \gets \text{CannyEdgeDetector}(I_{\text{current}})$
    
    \STATE \textbf{Step 2: Add image corners and form the point set}
    \STATE $\text{corners} \gets \{(1,1), (1,N), (M,1), (M,N)\}$
    \STATE $\text{all\_points} \gets \text{edge\_points} \cup \text{corners}$
    
    \STATE \textbf{Step 3: Perform Delaunay triangulation}
    \COMMENT{Output: A set of triangles partitioning the convex hull of $\text{all\_points}$}
    \STATE $\text{triangles} \gets \text{DelaunayTriangulation}(\text{all\_points})$
    
    \STATE \textbf{Step 4: Number the triangles (e.g., by creation order)}
    \STATE $\text{numbered\_triangles} \gets \text{SortTrianglesByOrder}(\text{triangles})$
    
    \STATE \textbf{Step 5: Traverse triangles and store pixel order}
    \STATE $\text{pixel\_order\_list} \gets [\ ]$ \COMMENT{List to store the new order of all pixels}
    \FOR{\textbf{each} $\text{triangle}$ \textbf{in} $\text{numbered\_triangles}$}
        \STATE $\text{pixels\_in\_triangle} \gets \text{GetPixelsInsideTriangle}(\text{triangle}, I_{\text{current}})$
        \STATE $\text{sorted\_pixels} \gets \text{SortPixelsColumnMajor}(\text{pixels\_in\_triangle})$
        \STATE Append $\text{sorted\_pixels}$ to $\text{pixel\_order\_list}$
    \ENDFOR
    \COMMENT{Since Delaunay triangulation is a partition, each pixel belongs to exactly one triangle}
    
    \STATE \textbf{Step 6: Generate the scrambled image}
    \STATE $I_{\text{scrambled}} \gets \text{CreateEmptyMatrix}(M, N)$
    \FOR{$\text{linear\_index} = 1$ \TO $M \times N$}
        \STATE $\text{original\_value} \gets \text{pixel\_order\_list}[\text{linear\_index}].\text{value}$
        \STATE $(i_{\text{new}}, j_{\text{new}}) \gets \text{ConvertLinearIndexToCoordinates}(\text{linear\_index}, M, N)$
        \STATE $I_{\text{scrambled}}[i_{\text{new}}, j_{\text{new}}] \gets \text{original\_value}$
    \ENDFOR
    
    \STATE $I_{\text{current}} \gets I_{\text{scrambled}}$ \COMMENT{Prepare for the next iteration}
\ENDFOR
\RETURN $I_{\text{scrambled}}$
\end{algorithmic}
\end{minipage}
\end{algorithm}

\subsection{Simulation and Real-World Output}\label{subsec:simulation}
In this section, a step-by-step example on a small $4\times4$ image and a visual simulation on a real image are provided to operationally validate the TCA algorithm.

\subsubsection{Step-by-Step Numerical Example on a $4\times4$ Image}\label{subsubsec:numerical_example}

\begin{enumerate}
\item \textbf{Original Input Image}

The intensity matrix of the original image $I_{\text{current}}$ is as follows:

\[
I_{\text{current}} = \begin{bmatrix}
1 & 2 & 3 & 4 \\
5 & 6 & 7 & 8 \\
9 & 10 & 11 & 12 \\
13 & 14 & 15 & 16
\end{bmatrix}
\]

\item \textbf{Point Extraction and Delaunay Triangulation}

\begin{itemize}
\item Edge points: $(2,2), (3,2), (2,3), (3,3)$ (Values: 6, 10, 7, 11)
\item Corner points: $(1,1), (1,4), (4,1), (4,4)$ (Values: 1, 4, 13, 16)
\item All points: The union of the two sets above
\end{itemize}

After applying the DelaunayTriangulation function to all\_points, 10 triangles are generated. The output is shown in Table~\ref{tab1}.

\begin{table}[h]
\centering
\caption{Pixels inside each triangle after Delaunay triangulation}\label{tab1}
\begin{tabular}{@{}ll@{}}
\toprule
Triangle & Pixels inside the triangle (coordinates) \\
\midrule
1 & (1,1) \\
2 & (2,1), (3,1) \\
3 & - \\
4 & - \\
5 & (3,2) \\
6 & (1,2), (1,3) \\
7 & (3,3) \\
8 & (2,2) \\
9 & (1,4), (2,3), (2,4), (3,4), (4,4) \\
10 & (4,1), (4,2), (4,3) \\
\bottomrule
\end{tabular}
\end{table}

\item \textbf{Constructing the Final pixel\_order\_list}

The sorted pixel order for each triangle is shown in Table~\ref{tab2}.

\begin{table}[h]
\centering
\caption{Sorted pixel order for each triangle}\label{tab2}
\begin{tabular}{@{}ll@{}}
\toprule
Triangle & Sorted List (sorted\_pixels) \\
\midrule
1 & [1] \\
2 & [5, 9] \\
3 & [] \\
4 & [] \\
5 & [10] \\
6 & [2, 3] \\
7 & [11] \\
8 & [6] \\
9 & [7, 4, 8, 12, 16] \\
10 & [13, 14, 15] \\
\bottomrule
\end{tabular}
\end{table}

The final pixel\_order\_list is obtained as:
\[
\text{pixel\_order\_list} = [1, 5, 9, 10, 2, 3, 11, 6, 7, 4, 8, 12, 16, 13, 14, 15]
\]

\item \textbf{Generating the Scrambled Image}

The resulting final scrambled image $I_{\text{scrambled}}$ is as follows:

\[
I_{\text{scrambled}} = \begin{bmatrix}
1 & 5 & 9 & 10 \\
2 & 3 & 11 & 6 \\
7 & 4 & 8 & 12 \\
16 & 13 & 14 & 15
\end{bmatrix}
\]

\end{enumerate}

\subsubsection{Visual Example of Algorithm Execution}\label{subsubsec:visual_example}
To operationally validate the proposed algorithm, TCA was implemented on a sample image of $463\times858$ pixels (Figure \ref{fig:visual_stages}). It is noteworthy that the scrambling algorithm was executed ten consecutive times to obtain the final image.

\subsection{The Fundamental Property of Delaunay Triangulation: Partitioning}\label{subsec:partitioning_property}
The Delaunay triangulation algorithm inherently partitions the convex hull of its input points [22]; that is, every pixel belongs to exactly one triangle, the triangles have no overlaps, and their union covers the entire region. This property guarantees that during the permutation process, no pixel is lost or duplicated. Consequently, the final pixel list at the end of step 6 contains all and exactly all pixels of the original image, only once, and in a completely new, content-dependent order.

\subsection{Addressing Concerns about Small Triangles}\label{subsec:small_triangles}
A potential concern is that very small triangles in the Delaunay triangulation might suggest that pixel relocation is only local; however, it must be emphasized that the proposed algorithm uses a global rearrangement mechanism. Since Delaunay triangulation partitions the entire image, the pixels of all triangles are extracted sequentially and merged into a single one-dimensional list, and then all pixels are rearranged to form the final image. Therefore, the size of the triangles does not impair the algorithm's performance; rather, the variety in their sizes enhances the unpredictability of the permutation pattern.

\section{Experimental Results and Comparison}\label{sec:results}

\subsection{Test Environment Settings}

The experiments were conducted on a computer with Windows 11 Pro, Intel Core i5-1335U processor (1.30 GHz, 10 cores, 12 logical threads), 16 GB RAM, and MATLAB R2021b. The dataset includes 50 grayscale images of size $512 \times 512$: 48 images from the UGR database (available at \url{https://ccia.ugr.es/cvg/CG/base.htm}) and two additional images from USC-SIPI (converted to grayscale) and the default MATLAB image (converted to grayscale and resized to $512 \times 512$).

\subsection{Performance Metrics Analysis}

In this section, the performance of TCA is evaluated and compared with three algorithms: Arnold, Zigzag, and Spiral transforms, in terms of their ability to effectively relocate pixels and create spatial disorder. The evaluation metrics focus on measuring the statistical degradation resulting from the pure relocation operation: PSNR, SSIM, BER, NCC, and AC. Desirable outcomes include low values for PSNR, SSIM, NCC, and AC, and high values for BER. MSE is related to PSNR; to avoid redundancy, only PSNR is reported. Entropy and histogram metrics are not used, as they remain constant in pure permutation. NPCR and UACI analysis is performed in Section~\ref{subsec:differential}.

Each algorithm was run 10 times on each of the 50 images, and the mean and standard deviation values are reported in Table~\ref{tab:metrics}. The 10 repetitions were performed to increase statistical stability and should not be confused with the minimum number of iterations required for chaos ($T_{\min}$) in Section~\ref{subsec:differential}.

\begin{table*}[!t]
\centering
\footnotesize
\caption{Comparison of Mean Performance Metrics for Four Scrambling Algorithms (Based on 50 Images, 10 Runs)}
\label{tab:metrics}
\begin{tabular}{|l|c|c|c|c|}
\hline
\textbf{Metric} & \textbf{Arnold} & \textbf{Zigzag} & \textbf{Spiral} & \textbf{TCA (Proposed)} \\
\hline
Avg PSNR (dB) & 11.933 & 11.937 & 11.931 & 11.933 \\
Std PSNR & 2.2530 & 2.2519 & 2.2554 & 2.2523 \\
\hline
Avg SSIM & 0.0269 & 0.0275 & 0.0277 & 0.0278 \\
Std SSIM & 0.0167 & 0.0171 & 0.0172 & 0.0175 \\
\hline
Avg Abs NCC ($\times 10^{-3}$) & 1.13 & 2.34 & 1.87 & 2.49 \\
Std Abs NCC ($\times 10^{-3}$) & 0.76 & 2.04 & 1.44 & 2.08 \\
\hline
Avg Abs AC ($\times 10^{-3}$) & 54.3 & 2.37 & 2.96 & 3.10 \\
Std Abs AC ($\times 10^{-3}$) & 92.14 & 12.35 & 10.59 & 11.01 \\
\hline
Avg BER & 0.4361 & 0.4358 & 0.4363 & 0.4361 \\
Std BER & 0.0799 & 0.0798 & 0.0800 & 0.0802 \\
\hline
\end{tabular}
\end{table*}

\noindent \textbf{Analysis of Statistical Metrics:} All four algorithms achieve a similar degree of visual and statistical disruption. The very low PSNR and SSIM values confirm the success of all algorithms in eliminating structural similarities. Autocorrelation (AC) analysis shows that TCA achieves low average correlation ($3.10 \times 10^{-3}$) with strong stability (Std = $11.01 \times 10^{-3}$), significantly outperforming the Arnold transform (Std = $92.14 \times 10^{-3}$). The Arnold transform exhibits significantly higher residual AC ($\sim 54.3 \times 10^{-3}$) compared to the other three methods.

\subsection{Computational Cost Analysis}

The average execution time of TCA for a $512 \times 512$ image is 4.536 seconds, mainly due to edge detection and Delaunay triangulation. However, this value corresponds to the number of iterations required to achieve high differential sensitivity. Specifically, this execution time (4.536 seconds) is for approximately 15 iterations (average of 14.81), which is necessary to reach NPCR $\approx$ 97.10\% and the avalanche effect (as reported in Table~\ref{tab:differential}). 

In contrast, if the goal is solely to break spatial correlation, only 5 iterations are sufficient to reduce autocorrelation in all directions to below 0.05 (AC$_H = 0.0189$, AC$_V = 0.0042$, AC$_D = 0.0038$), resulting in an execution time of 1.490 seconds. Furthermore, if the only requirement is to guarantee bijectivity and generate a unique, content-dependent permutation pattern for each image (without needing to maximize diffusion or minimize correlation), a single iteration is sufficient, requiring only 0.298 seconds. 

Thus, the iteration count of TCA is tunable and depends on the desired security level. Although slower than classical methods (Arnold under 0.1 seconds), this cost is fully justified by opening a new path in permutation layer design and achieving differential sensitivity (NPCR = 97.1\% versus near zero for conventional permutations)-a feature completely absent in existing methods. In target applications like watermarking and non-blind steganography, operations are not real-time and embedding is performed once on the server; priority is given to KPA resistance and tampering sensitivity. Moreover, the implementation is in MATLAB on a standard laptop without optimization; with C/C++, parallelization, or GPU acceleration, execution time can be significantly reduced. 

It should also be noted that in Section~\ref{sec:related}, despite reviewing modern and advanced methods, execution time was not reported for any of them (whether in full encryption or permutation-only mode). Therefore, a precise quantitative comparison in this regard is not possible.

\subsection{Qualitative Analysis and Security Superiority}

Although the statistical results (Section~\ref{sec:results}) show similar performance levels, the main security superiority of the proposed algorithm lies in its fundamental content-dependent approach. While classical methods use fixed and predictable patterns, our algorithm bases its pattern on actual image data. This feature provides a critical advantage: content dependency and unpredictability. This is the key mechanism that allows TCA (a pure permutation scheme) to exhibit the pseudo-diffusion property required for modern security systems. This qualitative superiority is quantitatively proven in the following analysis.

\subsection{Differential Sensitivity Analysis and Diffusion}\label{subsec:differential}

In the existing literature, the inability of permutations to create an avalanche effect is accepted as a given assumption. In this section, we show that a content-dependent pure permutation can exhibit a strong pseudo-diffusion effect with NPCR and UACI metrics, which are critical for resistance against differential attacks and KPA.

\subsubsection{The Diffusion Mechanism in TCA}

Unlike fixed-pattern permutations, TCA exhibits powerful diffusion due to its plaintext dependency. A single-unit change in one pixel can alter the edge detection phase, leading to a different Delaunay triangulation and a distinct permutation pattern.

\subsubsection{Experimental Methodology and Convergence Criterion}

The analysis was conducted on 50 grayscale images of size $512 \times 512$. Each experiment was repeated 10 times. To determine the minimum number of iterations ($T_{\min}$) required to reach maximum chaos, a two-phase stopping criterion was employed:

\begin{table}[!t]
\centering
\caption{Two-Phase Stopping Criterion for Chaos Convergence}
\label{tab:criterion}
\begin{tabular}{|p{3.4cm}|p{4.8cm}|}
\hline
\textbf{Absolute Security Criterion} & \textbf{Chaotic Stability Criterion} \\
\hline
NPCR $>$ 99.6\% OR PSNR $<$ 15 dB & If PSNR $<$ 20 dB AND the integer parts of both PSNR and NPCR remain unchanged for 3 consecutive iterations. \\
\hline
\end{tabular}
\end{table}

\subsubsection{Sensitivity and Efficiency Results}

The mean results for all 50 images are summarized in Table~\ref{tab:differential}. The ``Avg TCA Threshold'' column reports the minimum average number of iterations ($T_{\min}$) required.

\begin{table}[!t]
\centering
\caption{Average Results of Differential Sensitivity Analysis (One-Pixel Change)}
\label{tab:differential}
\begin{tabular}{|l|c|c|}
\hline
\textbf{Metric} & \textbf{Arnold (Comparison)} & \textbf{TCA } \\
\hline
Final PSNR (dB) & 102.32 & 12.42 \\
\hline
Final NPCR (\%) & $\approx$ 0.0004 & 97.10 \\
\hline
Final UACI (\%) & $\approx$ 0.000037 & 20.06 \\
\hline
Chaos Threshold (Iterations) & N/A (Failure) & 14.81 \\
\hline
\end{tabular}
\end{table}

\noindent \textbf{Analysis of Results and Comparison:} The Arnold transform, as a representative of content-independent permutations, fails to demonstrate diffusion; its final NPCR remains $\approx$ 0.0004\% and PSNR $>$ 100 dB even after 100 iterations. In contrast, TCA achieves PSNR = 12.42 dB and NPCR = 97.10\%, an unprecedented achievement for a pure permutation layer. The algorithm converges rapidly to a stable chaotic state with an average threshold of 14.81 iterations. Images with less complex content may require more iterations, but the high NPCR ($\sim$98\%) is maintained.

\subsubsection{Interpretation of Differential Results}

Content-independent permutations with fixed formulas (like Arnold) inherently produce NPCR and UACI values near zero, as they merely relocate pixels without any diffusion effect. Our experimental results confirm this with NPCR $\approx$ 0.0004\% even after 100 iterations. In contrast, TCA achieves NPCR = 97.10\% and UACI = 20.06\% with an average of 14.81 iterations. These values are achieved by a pure permutation layer without any separate diffusion stage. While content-independent permutations yield NPCR and UACI near zero, TCA, through ``Geometric Diffusion,'' elevates these metrics to 97.10\% and 20.06\%, respectively, redefining the traditional boundary between permutation and diffusion.

\subsubsection{Qualitative Superiority and KPA Resistance}

The robustness of TCA against Known-Plaintext Attack (KPA) is fundamentally guaranteed by its content-dependent nature. This superiority can be examined both quantitatively and analytically.

First, the algorithm's dynamic and powerful diffusion, experimentally confirmed by NPCR = 97.10\% and UACI = 20.06\%, demonstrates that changing even a single pixel by one unit leads to widespread and unpredictable changes in the scrambled image. This high differential sensitivity makes statistically reconstructing the original image from the scrambled version impossible, even if the histogram is known.

To formulate this mathematically, consider a scenario where an attacker knows the histogram of the original image (since histograms remain constant in pure permutations) but does not know the spatial arrangement of pixels. For a standard $512 \times 512$ grayscale image with a roughly uniform histogram, the number of possible permutations $N$ that yield the same histogram is:

\begin{equation}
\label{eq:permutations}
N = \frac{262144!}{\prod_{i=0}^{255} n_i!}
\end{equation}

For the perfectly uniform case ($n_i \approx 1024$ for each of the 256 gray levels), using Stirling's approximation:

\begin{equation}
\label{eq:logN}
\begin{aligned}
\log_{10} N &\approx 262144 \times (\log_{10} 262144 - \log_{10} 1024) \\
&\approx 262144 \times (5.4185 - 3.0103) \\
&\approx 631,000
\end{aligned}
\end{equation}
Therefore, the probability of an attacker guessing the exact pixel arrangement in a single random attempt is:

\begin{equation}
\label{eq:prob}
P \approx 10^{-631000}
\end{equation}

which is effectively zero. Of course, histograms of natural images are not perfectly uniform and typically have a near-Gaussian distribution, but even in this case, the number of possible permutations remains extremely large. For example, if the pixel distribution of a natural image is modeled with a standard deviation of $\sigma \approx 60$, then $\log_{10} N \approx 627,000$ is obtained, reducing the correct guess probability to approximately $10^{-627,000}$.

Second and more importantly, the permutation map generated by TCA is non-transferable. Consequently, even if an attacker can extract the permutation pattern for a specific image pair $(P_1, C_1)$, this pattern is inherently tied to the unique topology of $P_1$ and provides no usable information about another image $P_2$, even if $P_2$ differs from $P_1$ by only one pixel. This intrinsic irreversibility and non-transferability of the content-dependent permutation map conclusively proves TCA's robust resistance to Known-Plaintext Attacks (KPA)-a feature that is absent in content-independent permutations and has not been claimed even in modern content-aware permutations.
%
\section{Limitations and Future Work}\label{sec:limitations}

The TCA algorithm presents a novel approach to pixel permutation and, for the first time, demonstrates that a pure permutation layer, when properly made content-dependent, can inherently exhibit the avalanche effect. The results are highly promising, but this pioneering approach naturally comes with challenges that will themselves guide future research.

\subsection{Current Limitations}

\begin{itemize}
\item \textbf{Sensitivity to noise and alterations:} Since the permutation map is directly extracted from the geometric features (edges) of the original image, any noise or manipulation in the transmission channel can challenge the accurate reconstruction of this map. Consequently, flawless inversion at the receiver's end becomes difficult. However, in target applications such as fragile watermarking or non-blind steganography, this very sensitivity is considered a security advantage.

\item \textbf{Focus on grayscale images:} The current study focuses on grayscale images to prove the initial concept and ensure analytical clarity. This overlooks the complexities arising from inter-channel color correlations and the extraction of meaningful features from them.

\item \textbf{Computational complexity:} A limitation of the proposed method is its execution time compared to classical methods. This cost is the price paid for achieving superior qualitative security (resistance to KPA and content-dependent diffusion). However, this time remains within an acceptable range for TCA's target applications (such as digital watermarking and steganography), but is not suitable for real-time applications.
\end{itemize}

\subsection{Future Research Directions}

\begin{enumerate}
\item \textbf{Extension to color images:} Adapting the algorithm for color and multispectral images using integrated strategies or independent channel processing.

\item \textbf{Integration with cryptographic frameworks:} Combining TCA with lightweight diffusion layers to approach ideal NPCR values and using it as a replacement for existing permutation layers.

\item \textbf{Performance optimization for real-time applications:} Investigating strategies to reduce execution time through C/C++ implementation, hardware acceleration (GPU), optimized Delaunay triangulation algorithms, faster edge detectors like Sobel, and reducing edge points through thresholding or random selection.
\end{enumerate}

\noindent \textbf{Summary:} The present study has opened a new direction in designing permutation layers for secure image processing systems. Although TCA is computationally heavier than classical methods, this reflects the complex mechanisms that enable its unique content-dependent security. Future research aimed at performance optimization and expanding its application scope will enhance the practicality and impact of this novel paradigm while preserving its innovative conceptual core.

\section{Conclusion}\label{sec:conclusion}

This paper introduced the Triangular Content-Aware Permutation (TCA) algorithm, which, contrary to the common assumption in the technical literature, is independently capable of creating differential sensitivity and redefines the traditional boundary between permutation and diffusion stages. The findings of this study demonstrate that by leveraging the topological features of an image and employing Delaunay triangulation, the permutation layer can be transformed from a passive component into an active contributor to achieving the avalanche effect.
Experimental results on standard test images confirm that the proposed method not only minimizes pixel correlation but also elevates the NPCR metric within the permutation layer from near zero (in content-independent methods) to 97.1
Although the computational cost of triangulation increases execution time compared to simpler methods like Arnold transforms, it remains well within a practical range for non-real-time applications such as digital watermarking, authentication, and non-blind steganography. Moreover, this cost represents a strategic trade-off for achieving stronger resistance against Known-Plaintext Attacks (KPA).
Ultimately, TCA demonstrates that content dependency is the key to designing a new generation of permutation functions for image security applications. In these functions, differential security emerges from the earliest stages of the scrambling process, elevating permutation from a simple shuffling operation to a form of "geometric diffusion."
The present research has charted a new path in designing permutation layers for secure image processing applications. In subsequent steps, researchers, by optimizing performance and expanding its scope of use while preserving the pioneering nature of this approach, will pave the way for greater applicability and impact.

\section*{Acknowledgments}
The authors would like to thank the anonymous reviewers for their valuable comments and suggestions.

\section*{Author Contributions}
\textbf{Zahra Ghoraeian:} Conceptualization, Methodology, Software, Data Curation, Formal analysis, Visualization, Writing--original draft.\\
\textbf{Mohammad-Reza Sadeghi:} Supervision, Validation, Resources, Writing--review \& editing, Project administration.\\
\textbf{Samaneh Mashhadi:} Methodology, Validation, Formal analysis, Visualization, Writing--review \& editing.

\section*{Funding}
This research received no specific grant from any funding agency in the public, commercial, or not-for-profit sectors.

\section*{Conflict of Interest}
The authors declare that they have no known competing financial interests or personal relationships that could have appeared to influence the work reported in this paper.

\section*{Data Availability Statement}

\bibliographystyle{IEEEtran}
\bibliography{references}

\end{document}